# An Attention-Guided Graph Neural Network Surrogate for Boundary Integral Computation in Physical Field Reconstruction

[Author: RongZe Gao[a], YuWei Wu[a], XinRan Ji[a] , XingYue Ren[a]* ]

[[a] Affiliation: The School of Civil and Architectural Engineering, Hainan University, HaiKou, 570000, Chain.]

[* Corresponding author: XingYue Ren; E-mail: [renxny@126.com]]

[ORCID iDs ): RongZe Gao — [0009-0006-2617-6735]

## Abstract

In practical engineering and scientific computing, a real physical system can be reduced to a boundary value problem, and the adequate boundary information to determine the interior spatial distribution and evolutionary state of the physical field can be solved with linear equations, alternatively, the the real physical problem can be treated with boundary problems with in adequate information problems in field of machine learning. This paper proposes a deep learning framework for the recovery of missing boundary information using an attention-guided graph neural network (GNN) surrogate. The framework extracts features from the physical information on the discretized boundary into a latent space, and the missing boundary quantities are then predicted by the network. Thus, the nonlinear mapping from known boundary quantities to missing boundary quantities is learned using a variational autoencoder (VAE) architecture, while an attention mechanism is incorporated to better model the correlations among different types of boundary information. Furthermore, the interior physical field can be reconstructed using the conventional boundary element method (BEM) integral formulation. The effectiveness of the GNN-BEM framework is evaluated using numerical examples of the two-dimensional Laplace and Poisson equations, which demonstrate good performance in reconstruction accuracy, local detail preservation, and model stability. The results show that GNN-BEM provides a feasible computational framework for boundary information recovery and fast interior physical field reconstruction for partial differential equations under complex boundary conditions.

## 1 Introduction

The application of artificial intelligence (AI) continues to expand in engineering and scientific computing, particularly in the modeling and analysis of physical systems[1,2]. In this context, physics-informed neural networks (PINNs), which incorporate governing differential equations into neural-network training, have attracted increasing attention[3,4]. Existing studies generally focus on three directions: discovering governing equations from observational data, efficiently solving partial differential equations (PDEs)[5], and learning representations of solution operators that generalize across input conditions[6,7]. Several landmark studies illustrate these directions. For example, Raissi[8] proposed a data-driven method for identifying unknown nonlinear PDEs from observations and validated it on representative systems, including the Burgers and Navier–Stokes equations. Subsequently, Raissi et al.[3] developed the PINN framework, which incorporates governing equations, boundary conditions, and initial conditions into neural-network training to address forward and inverse PDE problems. Lu et al. [9] developed DeepONet to learn mappings from input functions to output solution functions, enabling fast prediction under varying input conditions. Brunton and Kutz [6] highlighted the use of learned solution-operator representations to improve[10], or potentially replace, traditional numerical algorithms as a promising direction for machine learning in PDE research.

For well-posed PDE boundary-value problems, the governing equations and prescribed boundary conditions jointly determine the physical field [11]. At the numerical level, practical difficulties mainly involve accurately representing boundary geometry and prescribed boundary conditions[12], capturing the coupling among different types of boundary physical quantities, and reconstructing complete boundary data from sparse boundary measurements[13,14]. Based on this understanding, the boundary element method (BEM) [15] provides a natural framework for modeling boundary physical quantities. By transforming a domain problem into boundary integral relations[16], BEM can restrict discretization to the boundary when a boundary-only formulation is available[17], thereby reducing the geometric dimension by one[18], for example, from three dimensions to two or from two dimensions to one[19]. This transformation concentrates the representation and coupling of boundary quantities at boundary nodes[20], allowing different boundary nodal quantities to be directly related while reducing the need to collect full-domain mesh-point information. As a result, this boundary-centered formulation may alleviate the learning burden[21] associated with dense full-domain sampling in PINN approaches [22].

Because BEM reformulates a PDE problem in terms of boundary integral relations, the geometric coordinates and physical quantities of boundary nodes can be represented jointly, aligning naturally with graph connectivity and the message-passing mechanism of graph neural networks (GNNs) [23,24]. As shown in Fig. 1 and Eq. (1), this study develops a deep learning framework that combines neural-network inference with the mathematical formulation of BEM[25,26]. Specifically, the coordinates and known physical quantities of boundary nodes are first represented as graph-node features[27], and an attention-guided GNN surrogate[28] then predicts the missing boundary quantities from the observed boundary data[29,30]. The completed boundary data are subsequently used to evaluate the interior physical field[31] through a Green's-function-based boundary integral equation [11,27]. Thus, this study treats the mapping from known to unknown boundary quantities[32] as a learnable nonlinear surrogate for the BEM boundary-computation stage, while retaining the conventional BEM integral formulation for interior field reconstruction[33]. Compared with PINN approaches that enforce PDE constraints at collocation points throughout the domain [12,13,25,26], the proposed boundary-centered strategy may reduce the training burden and improve computational efficiency in complex physical and engineering problems.Code availability

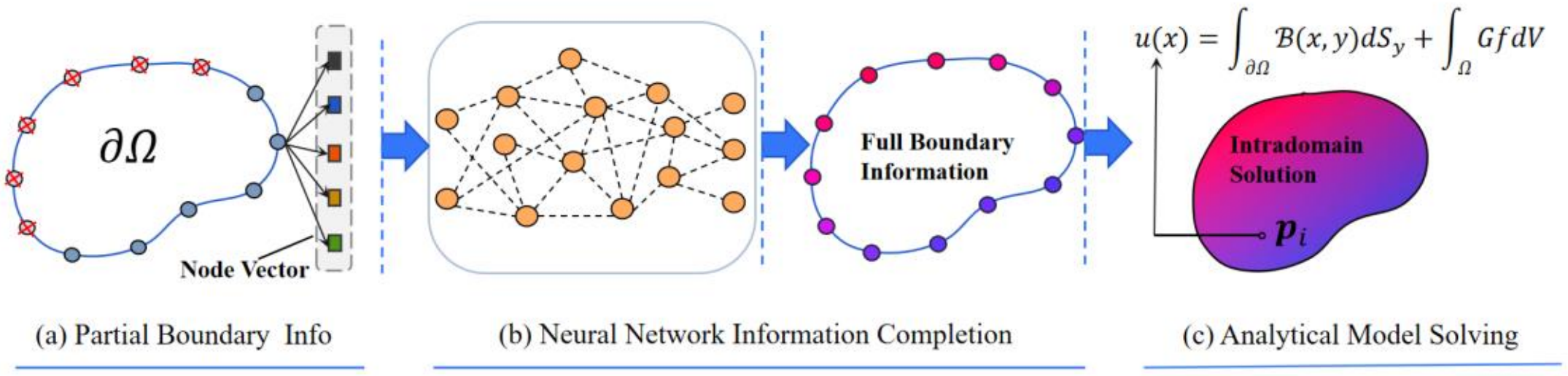

Fig.1 Architecture of the GNN-BEM model. (a) Partial boundary information and the corresponding graph representation of nodes; (b) a GNN-based variational auto-encoder network with the attention mechanism; (c) an integral solver for physical information in the domain（interior-domain analytical model, the integral formula for $\xi \in \Omega$ when using Equation (1)).

## 2. Methods

### 2.1 Boundary Integral and Surrogate Mapping

Applying Green's identity together with the fundamental solution of the governing equation yields the boundary integral equation (BIE):

$$c(\xi)u(\xi) + \int_{\partial\Omega} u(\eta)\frac{\partial G(\xi,\eta)}{\partial n(\eta)} d\Gamma(\eta) = \int_{\partial\Omega} q(\eta)G(\xi,\eta)d\Gamma(\eta), \quad \xi \in \Omega \cup \Gamma \tag{1}$$

where $\Omega \subset \mathbb{R}^d$ and $\Gamma = \partial\Omega$ denote the computational domain and its boundary, respectively. $u$ and $q$ represent the Dirichlet and Neumann boundary quantities defined on the boundary. $\xi$ and $\eta$ denote the boundary collocation and integration points, respectively. $G(\xi,\eta)$ denotes the Green's function associated with the governing equation, while $c(\xi)$ is the free-term coefficient determined by the local boundary geometry.

Following boundary discretization, the geometric and physical information associated with the boundary is represented by the discrete boundary state $S_h$:

$$S_h = \{s_j\}_{j=1}^{\mathrm{N}}, \quad s_j = \left[r_j, \quad n_j, \quad \Delta s_j, \quad f_{p,j}\right]_{j=1}^{N} \tag{2}$$

In Eq. (2), $N$ denotes the number of boundary nodes, and $s_j$ represents the state associated with the $j\!-\!th$ boundary node. This state includes the spatial coordinates, outward unit normal vector, associated boundary-element length, and source-term feature.

Given the governing differential operator, its fundamental solution, the boundary topology, and the discrete boundary state, the BEM matrix-assembly process can be abstracted as the following mapping:

$$(H_h, G_h) = \psi_h(\mathcal{P}, G_{\mathcal{P}}, \Gamma; S_h) \tag{3}$$

In Eq. (3), $\psi_{\mathrm{h}}$ denotes the BEM matrix-assembly operator. Applying this operator to the BIE in Eq. (1) yields the following algebraic boundary system:

$$H u - G q = 0 \tag{4}$$

For prescribed Dirichlet boundary values, Eq. (4) defines an explicit Dirichlet-to-Neumann mapping $\phi(\cdot)$, provided that $G$ is invertible, as Eq. (5):

$$q = G^{-1} H u = \phi_{H,G}(u) \tag{5}$$

The complete BEM boundary-computation mapping $\varphi(\cdot)$ accounts for both the boundary geometry and features encoded in the discrete boundary state $S_h$ and the prescribed $u$. An attention-guided GNN surrogate $N_\theta$ [35], parameterized by $\theta$, is therefore introduced to approximate $\varphi$, predicting the unknown $q$:

$$q = \varphi(S_h, u) \approx N_\theta(S_h, u) \tag{6}$$

**2.2 Boundary Graph Representation and Attention-Guided GNN Surrogate**

As illustrated in Fig. 2, the neural-network component of the GNN-BEM framework adopts a graph-attention-based variational autoencoder (VAE) architecture[36]. Following a modular design strategy[37], the encoder consists of two sequential encoding blocks with the same configuration, each containing a graph-attention message-passing layer[38], layer normalization[39], and dropout[40].The decoder comprises three stacked decoding blocks, each incorporating graph-attention messagepassing, layer normalization, GELU activation [41], and dropout. Two separate decoder output branches are employed to predict Dirichlet and Neumann boundary quantities, respectively, with the required predictions selected according to the prescribed boundary-condition types.

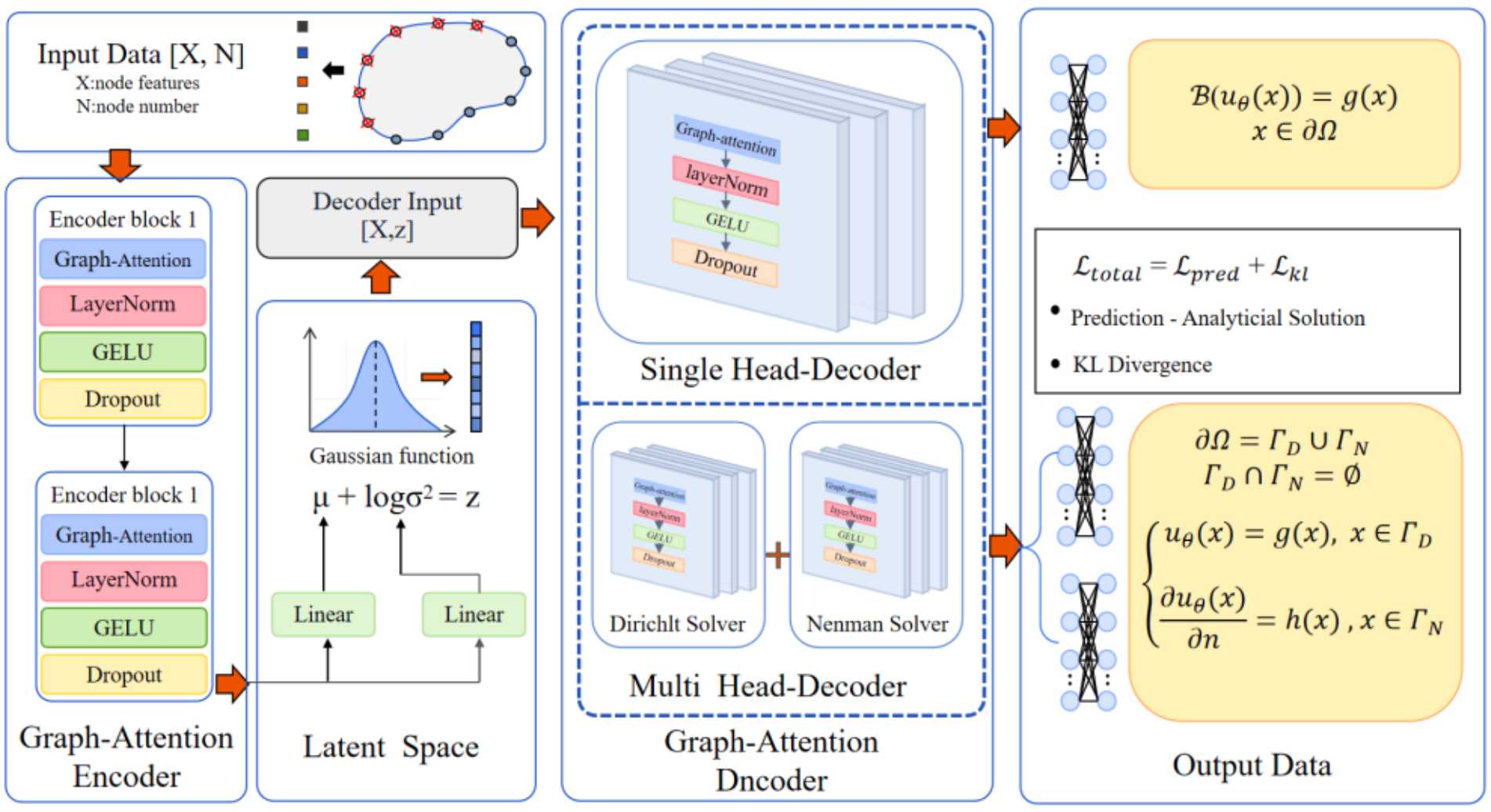

**Fig.2 Schematic diagram of the deep neural network architecture in the GNN-BEM framework**

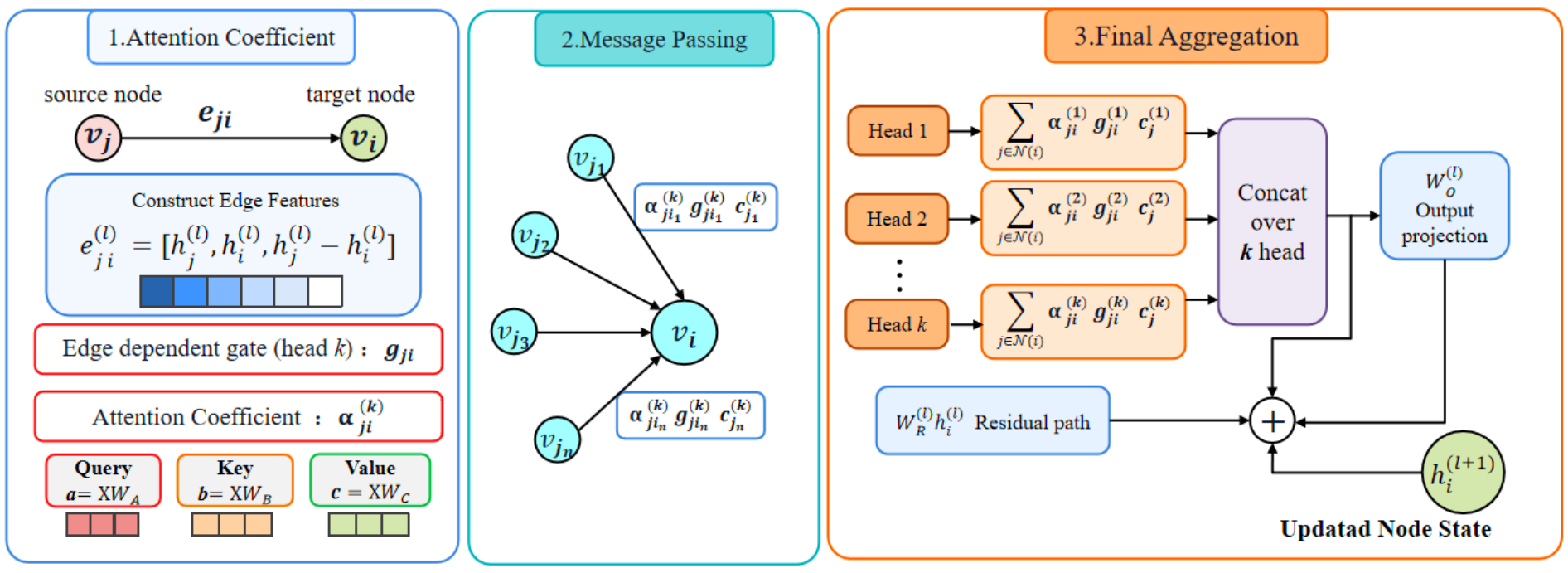

**Fig.3 Schematic of the graph-attention message-passing mechanism in the GNN-BEM framework.**

BEM and GNNs share a node-to-node interaction structure[42], motivating the representation of the discretized boundary as a fully connected directed graph:

$$\mathcal{G} = (\mathcal{V}, \mathcal{E}),\ \ \mathcal{V} = \{v_j, x_j\}_{j=1}^{N},\ \ \mathcal{E} = \{v_j, v_i\}_{j,i=1}^{N} \tag{7}$$

where $\mathcal{G}$ denotes the boundary graph, $\mathcal{V}$ is the set of boundary nodes, and $\mathcal{E}$ is the set of directed edges[43]. Each node $v_j$ corresponds to the $j-th$ discretized boundary point, while the directed edge $e_{ji}$ represents information transfer from the source node $v_j$ to the target node $v_i$.

The geometric and physical attributes of the boundary nodes are organized into an input feature matrix (IFM)[44], defined as:

$$x_j = [\ r_j, n_j, \Delta s_j, b_j^{known},\quad m_j, f_{p,j}] \tag{8}$$

where $b_j^{known}$ denotes the feature value associated with a boundary node, and $m_j$ represents the boundary-condition mask indicating the type of prescribed boundary quantity, (0, 1) corresponds to a prescribed Dirichlet value, whereas (1, 0) corresponds to a prescribed Neumann value.

As illustrated in Fig. 3, the graph-attention encoder consists of $\boldsymbol{l}$ message-passing layers [45]. The graph topology remains unchanged across layers, while the feature representation of each node is updated [46]. The initial node features and the directed relational features from $\boldsymbol{v_j}$ to $\boldsymbol{v_i}$ are defined as:

$$a_i^{(k)}=(W_A h_i^{(l)})^{(k)},\ b_j^{(k)}=(W_B h_j^{(l)})^{(k)},\ c_j^{(k)}=(W_C h_j^{(l)})^{(k)}$$
$$h_j^{(0)}=x_j,\ e_{ji}^{(l)}=[h_j^{(l)},h_i^{(l)},h_i^{(l)}-h_j^{(l)}] \tag{9}$$

Where $h_j^{(l)}$ denotes the feature representation of node $j$ at layer $l$. The pairwise relational feature incorporates the representations of both interacting nodes and explicitly encodes their feature difference[42].

For the *k-th* attention head, the attention coefficient[47] is expressed $\alpha_{ji}^{(k)}$ and the gate coefficient of the attention head $g_{ji}^{(k)} \in (0,1)$ as:

$$\begin{cases} \alpha_{ji}^{(k)} = Softmax_{j\to i}\left[\dfrac{(a_i^{(k)})^T(b_i^{(k)}))}{\sqrt{d_h}} + B^{(k)}(e_{ji}^{(l)})\right] \\ g_{ji} = \sigma\left[(W_{g,2}\, Gelu(W_{g,1}e_{ji}^{(l)} + b_{g,1}) + b_{g,2})\right], \qquad g_{ji} = [g_{ji}^{(1)}, g_{ji}^{(2)} \cdots g_{ji}^{(k)}] \end{cases} \tag{10}$$

In Eq.(10), $d_h$ denotes the feature dimension of a single attention head, $B^{(k)}(\cdot)$ denotes the learnable relational bias for the (k)−th attention head generated from the relational features. $g_{ji}$ is the edge-dependent gating vector from $v_j$ to $v_j$. $W_{g,1}$ and $W_{g,2}$ are the trainable weight matrices of the gating network, while $b_{g,1}$ and $b_{g,2}$ are the corresponding bias vectors.

Using the attention and gating coefficients defined in Eq. (10), the attention heads perform weighted aggregation of source-node representations for each target node[48]:

$$h_i^{(l+1)} = W_R^{(l)}\, h_i^{(l)} + W_O Concat_{k=1}^{K} \sum_{j\in\mathcal{N}(i)} \alpha_{ji}^{(k)} g_{ji}^{(k)} c_j^{(k)} \tag{11}$$

$N(i)$ denotes the set of source nodes that transmit information to $v_i$. $W_O^{(l)}$ is the output projection matrix that maps the aggregated and concatenated information from different attention heads onto the output feature space of the current graph-attention layer, whereas $W_R^{(l)}$ is the residual projection matrix used to preserve the current feature representation of the target node[37].

After two encoding blocks , the model constructs an independent latent distribution for each boundary node[49]:

$$\mu_i = W_\mu h_i^e + b_\mu, \qquad \log\,\sigma_i^2 = W_\sigma h_i^e + b_\sigma$$
$$z_i = \mu_i + exp(\frac{1}{2} log\ \sigma_i^2) \odot \epsilon, \qquad \epsilon_i \sim N(0, I) \tag{12}$$

To preserve the original boundary features during decoding, the latent representation is concatenated with the original node features:

$$h_i^{d,0} = [x_i, z_i], q_i = W_{out} D_\theta(\{h_j^{d,0}\}_{j=1}^{N_b}, \mathcal{E})_i + b_{out} \tag{13}$$

where $D_\theta(\cdot)$ represents the decoder. The network is trained using a reconstruction loss together with the regularization of the node-wise latent distributions[50]:

$$\mathcal{L} = \frac{1}{N_b} \sum_{i=1} \| y_i - \hat{y}_i \|_2^2 + \beta \left[ -\frac{1}{2} mean_{i,r} (1 + \log \sigma_{i,r}^2 - \mu_{i,r}^2 - exp(\log \sigma_{i,r}^2)) \right] \tag{14}$$

where $\beta$ controls the regularization strength of the latent space and is set to $10^{-3}$ in the present implementation.

## 3 Experimental results

To verify the feasibility of the proposed GNN-BEM framework, three two-dimensional benchmark cases are considered, including a Laplace equation with mixed boundary conditions, a Poisson equation with a polynomial source term, and a Poisson equation with a sinusoidal source term. The predicted boundary quantities are combined with the prescribed boundary values and further used in the BEM integral formulation to reconstruct the interior physical field. The accuracy of both boundary recovery and interior field reconstruction is evaluated by comparison with analytical solutions.

### 3.1 Two-Dimensional Laplace Equation with Mixed Boundary Conditions

The first example considers a source-free two-dimensional Laplace problem. The governing equation is

$$-\Delta u(x, y) = 0, \quad (x, y) \in \Omega, \tag{15}$$

The rectangular domain is $\Omega = (0, 2) \times (0, 1)$. Mixed boundary conditions are imposed: the Dirichlet values are $u = 1, 2$ on the left and right edges, respectively, the Neumann values are $\partial u / \partial n = 0, 0$ on the upper and lower edges. With this configuration, the analytical reference solution is：

$$u_{ref}(x,y) = 1 + \frac{x}{2} \quad (16)$$

The known boundary inputs are illustrated in Fig. 4, and the complete analytical boundary distributions are shown in Fig. 5. The Dirichlet quantity varies across the rectangle, whereas the Neumann quantity is zero on the horizontal edges and takes opposite signs on the two vertical edges. Together, these plots establish the reference boundary information against which the recovery results are evaluated.

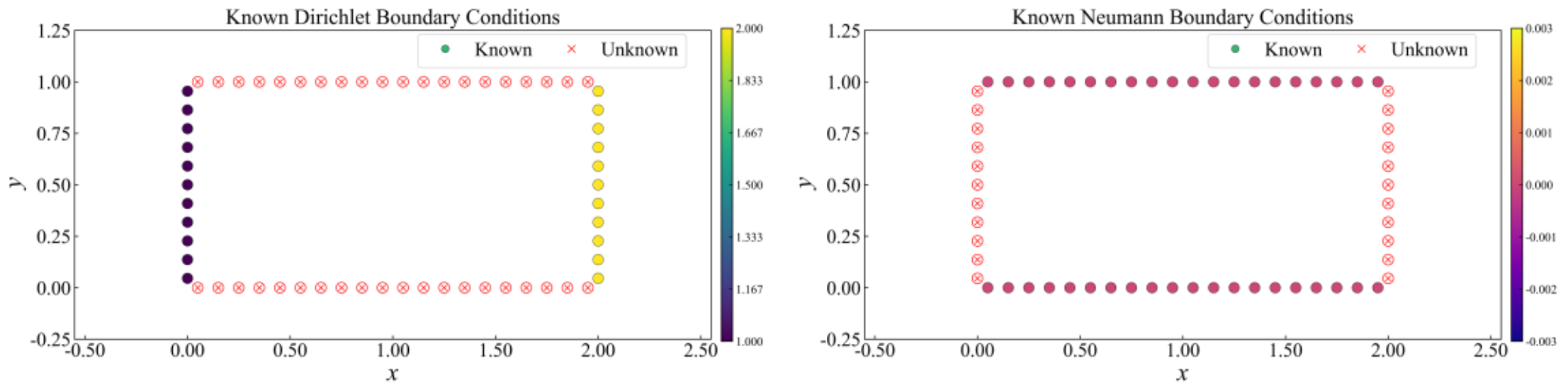

(a) (b)

**Fig. 4. Prescribed boundary information for the mixed-boundary Laplace benchmark:** (a) Dirichlet values on the left and right edges; (b) homogeneous Neumann values on the upper and lower edges. The remaining boundary quantities are treated as unknown.

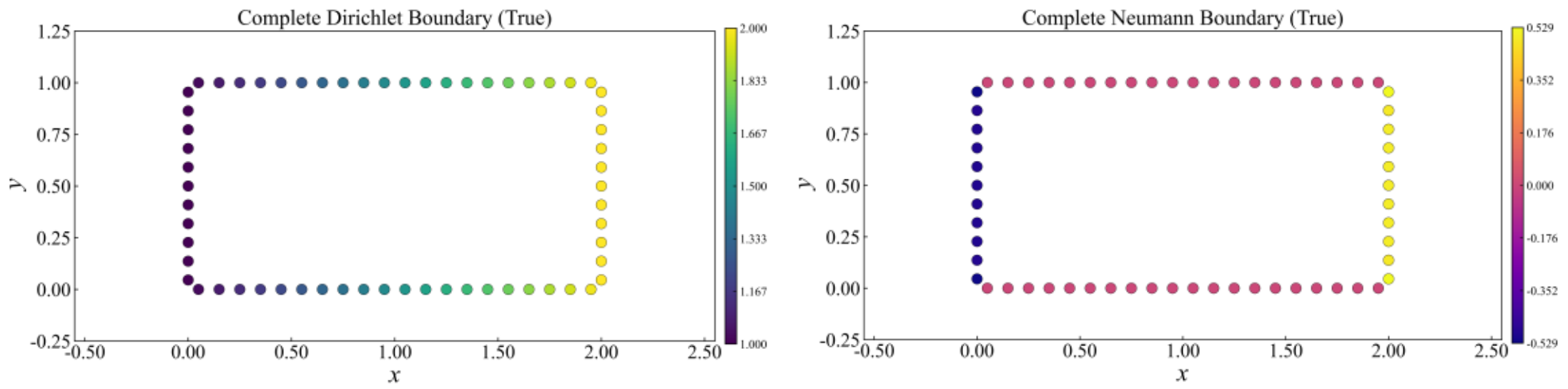

(a) (b)

**Fig. 5. Analytical boundary distributions for the Laplace benchmark:** (a) Dirichlet quantity u; (b) outward normal derivative q.

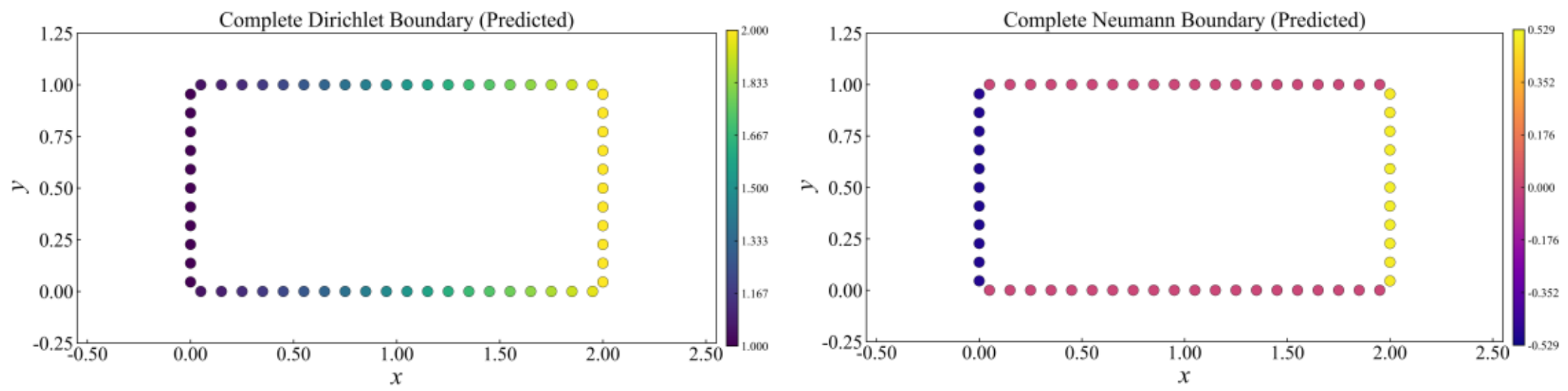

(a) (b)

**Fig. 6. Completed boundary distributions for the Laplace benchmark:** (a) Dirichlet quantity u; (b) Neumann quantity q. The visualizations combine prescribed values with GNN predictions at complementary boundary locations.

The corresponding boundary prediction errors are plotted in Fig. 7. The reported maximum absolute errors are 0.024 for the recovered Dirichlet quantity and 0.04 for the recovered Neumann quantity. Larger deviations are concentrated near the corners, where the boundary direction and outward normal change between adjacent edges. These local deviations do not alter the overall agreement of the predicted boundary distributions.

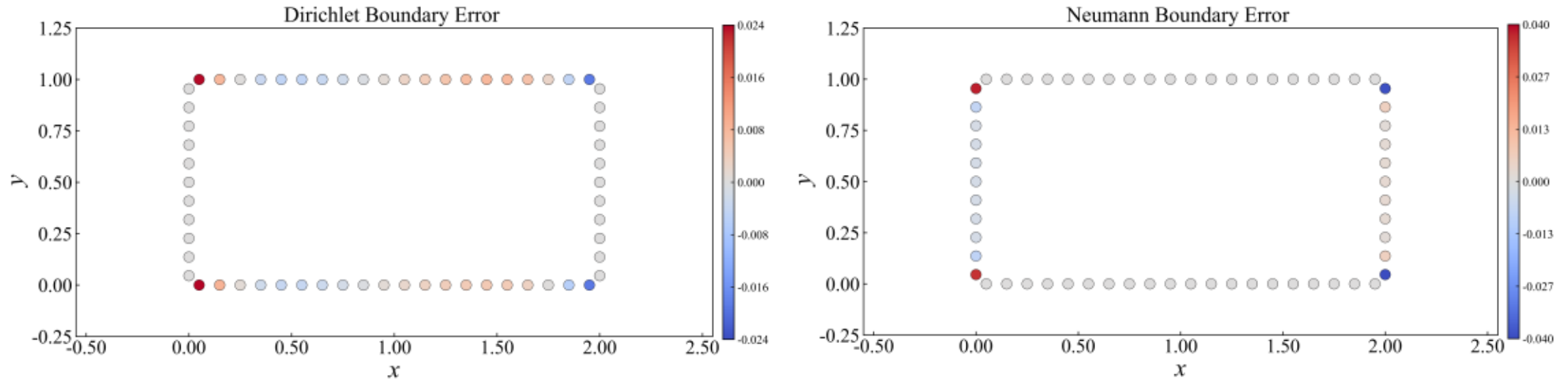

(a) (b)

**Fig. 7.** Boundary prediction errors for the Laplace benchmark: (a) error in Dirichlet values on the horizontal edges; (b) error in Neumann values on the vertical edges.

To examine the second stage of the framework, the completed boundary information is substituted into the BEM integral formulation. Figure 8 compares the resulting interior field with the reference BEM field computed from analytical boundary values. Their principal spatial distributions are similar, with a reported maximum absolute difference of 0.032. This observation supports the feasibility of using the recovered boundary information for interior-field reconstruction in the present source-free benchmark.

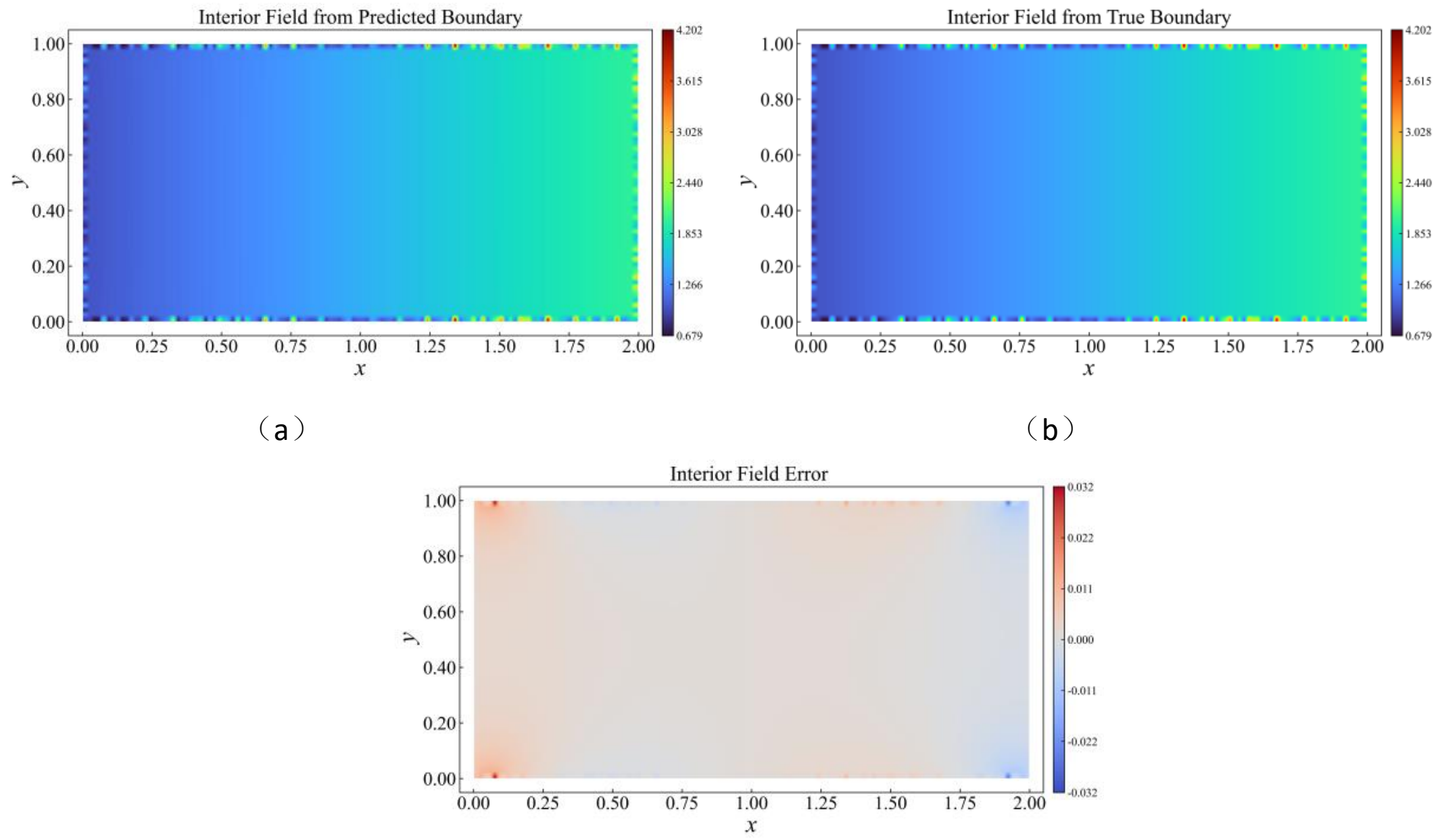

(c)

**Fig. 8. BEM interior-field reconstruction for the Laplace benchmark:** (a) field obtained from the completed boundary information; (b) reference field computed using analytical boundary data; (c) pointwise difference.

### 3.2 Two-Dimensional Poisson Equation with Different Source Terms

Two Poisson benchmarks are considered to extend the feasibility check to problems with non-zero source terms. Both satisfy the governing relation $-\Delta u = f$. A manufactured-solution procedure is adopted: an analytical reference field is prescribed and its source term is evaluated from $f = -\Delta u_{ref}$ .

### 3.2.1 Poisson Equation with a Polynomial Source Term

The first Poisson benchmark is defined on the unit disk $\Omega = (x, y): x^2 + y^2 < 1$. A polynomial analytical field is selected in the form

$$u_{ref}(x, y) = (1 - x^2 - y^2)(a + bx + cy) \quad (17)$$

where the coefficients in the reported test are $\boldsymbol{a} = 1.47$, $\boldsymbol{b} = -0.24$, and $\boldsymbol{c} = 0.85$. The factor $1 - x^2 - y^2$ yields zero Dirichlet values on the circular boundary. The corresponding source term, obtained by applying the governing differential operator to Eq. (17), is Eq. (18):

$$f(x, y) = 4a + 8bx + 8cy) \quad (18)$$

Figure 9 shows the prescribed Dirichlet inputs, the analytical Neumann distribution, the GNN prediction, and the pointwise prediction error. The analytical Neumann values vary smoothly around the circle, and the predicted values reproduce this overall trend. The reported maximum absolute boundary prediction error is approximately 0.05, indicating close agreement for this benchmark.

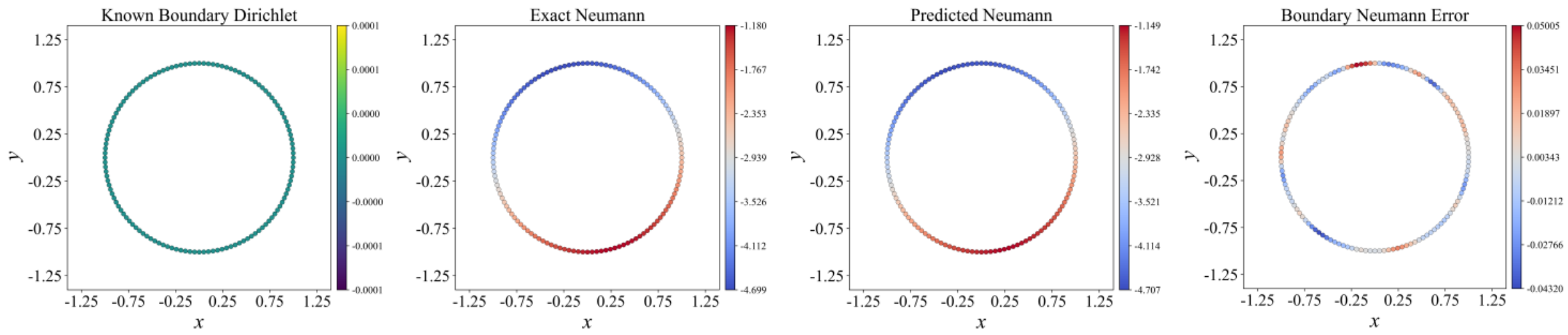

（a） （b） （c） （d）

**Fig. 9. Boundary results for the Poisson benchmark with a polynomial source term:** (a) prescribed Dirichlet values; (b) analytical Neumann values; (c) GNN-predicted Neumann values; (d) boundary prediction error

The interior-field results are compared in Fig. 10. The BEM-computed field has a spatial distribution similar to that of the analytical reference field, and the reported maximum absolute difference is 0.015.

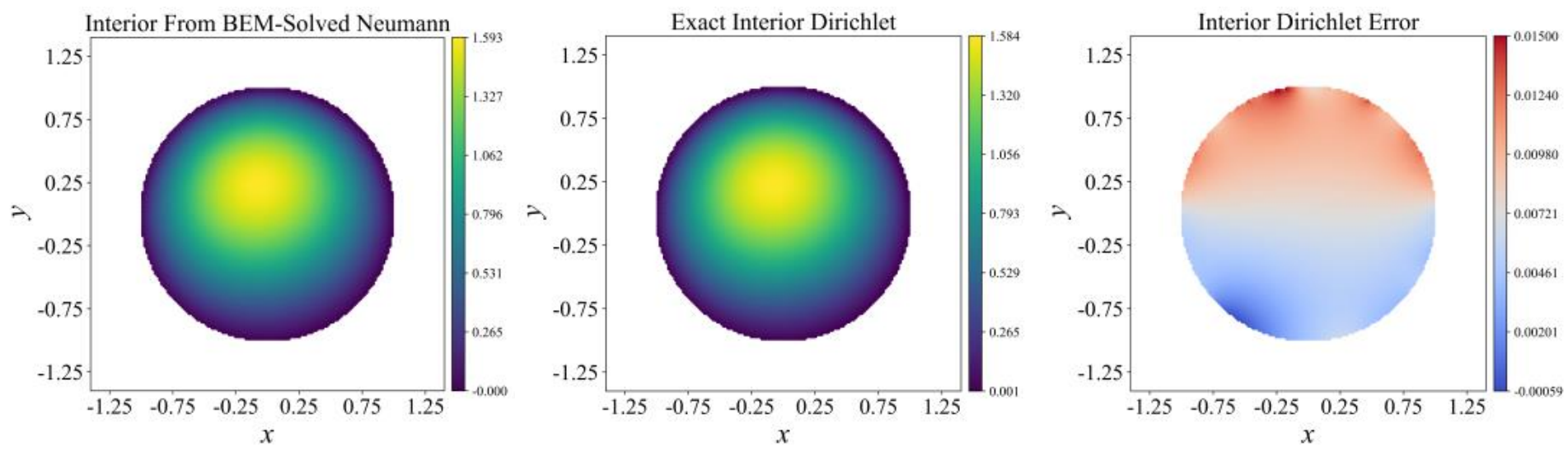

（a） （b） （c）

**Fig. 10. Interior-field comparison for the polynomial-source Poisson benchmark:** (a) BEM-computed field; (b) analytical reference field; (c) pointwise difference.

### 3.2.2 Poisson Equation with a Sinusoidal Source Term

The second Poisson benchmark is considered on the unit square $\Omega = (0,1) \times (0,1)$. To obtain a field with more pronounced spatial variations, the following sinusoidal analytical solution is prescribed:

$$u_{ref}(x,y) = A sin(m\pi x) \sin(n\pi y) \tag{19}$$

Here A is the amplitude and m and n are the spatial mode numbers. The reported parameters are $A = 1.77$, $m = 4$, and $n = 2$. This choice satisfies homogeneous Dirichlet conditions on the four edges. The corresponding source term is

$$f(x,y) = A\pi^2(m^2 + n^2) \sin(m\pi x) \sin(n\pi y) \tag{20}$$

Figure 11 compares the predicted Neumann boundary values with the analytical solution. The predicted results reproduce the main oscillatory variations of the analytical distribution, while the error plot reveals localized discrepancies. This indicates that the GNN surrogate can capture the principal characteristics of the missing boundary quantities under the sinusoidal source term.

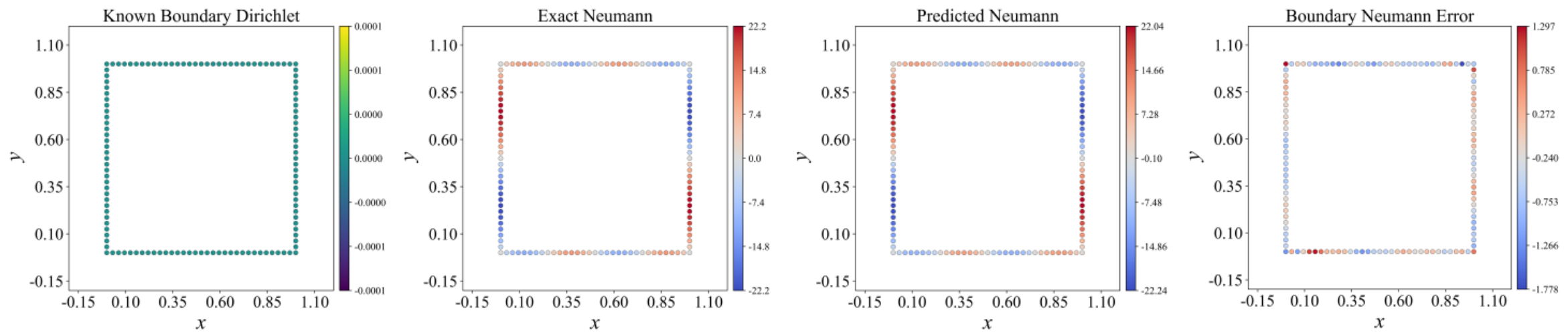

(a) (b) (c) (d)

**Fig. 11. Boundary results for the Poisson benchmark with a sinusoidal source term:** (a) prescribed Dirichlet values; (b) analytical Neumann values; (c) GNN-predicted Neumann values; (d) boundary prediction errors.

Figure 12 presents the interior field reconstructed using the predicted Neumann values and the corresponding BEM reference field. The reconstructed field shows good agreement with the reference in terms of the overall spatial distribution, although differences remain in certain regions. The maximum absolute difference is 0.2189, exceeding that of the polynomial-source case and indicating a larger reconstruction error for this benchmark.

In summary, the Laplace and Poisson benchmarks demonstrate the feasibility of the proposed two-stage procedure under the tested mixed-boundary and source-term settings. In all three cases, the recovered boundary quantities reproduce the main features of their analytical counterparts, and the subsequent BEM calculations provide interior fields with broadly consistent spatial distributions.

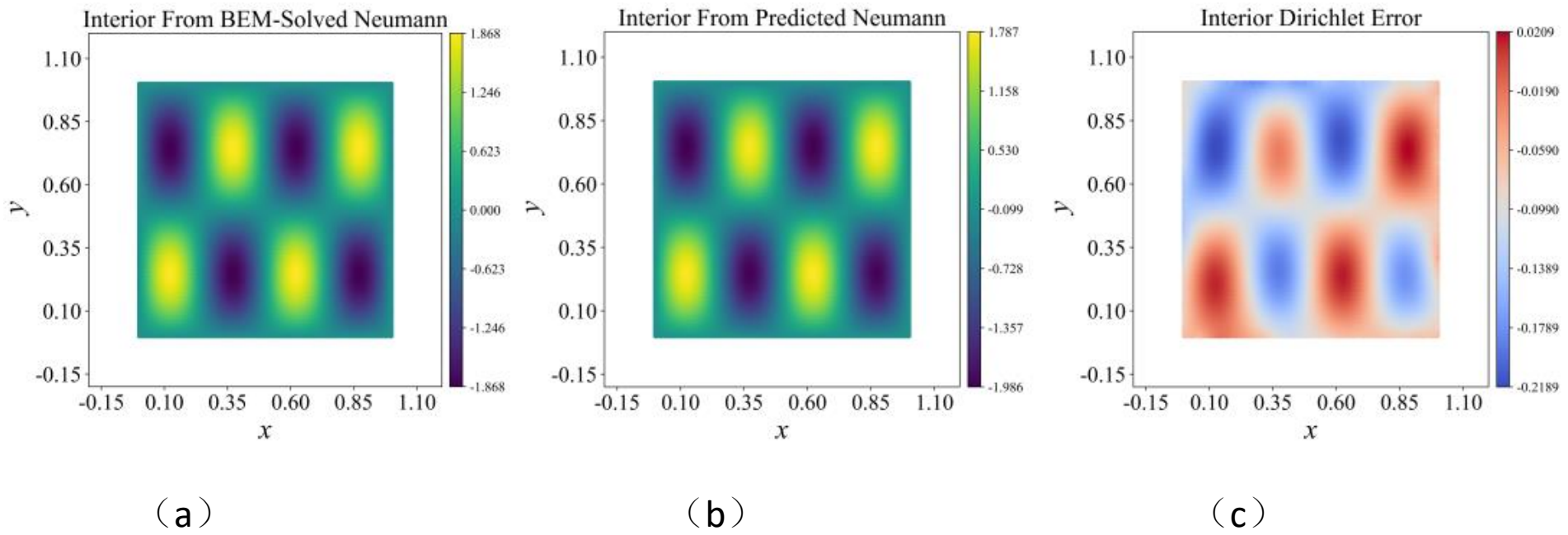

**Fig. 12. Interior-field comparison for the sinusoidal-source Poisson benchmark:** (a) BEM reconstruction using predicted Neumann values; (b) BEM comparison field obtained using BEM-solved Neumann values; (c) pointwise difference.

## 4 Conclusions

This study developed an attention-guided GNN–BEM framework for reconstructing physical fields from incomplete boundary data. The discretized boundary is represented as a graph, and a variational graph-attention encoder–decoder estimates the missing Dirichlet or Neumann quantities. The completed boundary data are then used in the conventional BEM integral formulation to reconstruct the interior field. Thus, learning approximates the boundary mapping while the integral solver is retained.

Regarding boundary recovery, the maximum absolute errors in the mixed-boundary Laplace case were 0.024 for Dirichlet data and 0.04 for Neumann data, while the polynomial-source Poisson case yielded a maximum absolute Neumann error of approximately 0.05. For the mixed-boundary Laplace problem and the polynomial- and sinusoidal-source Poisson problems, the reported maximum absolute interior-field differences were 0.032, 0.015, and 0.2189, respectively. These results support the feasibility of the method for the tested configurations, while the larger error in the sinusoidal case indicates problem-dependent accuracy. These benchmarks are limited to two-dimensional test cases and do not, by themselves, establish generalization to unseen geometries or a computational speedup. Generalization to other geometries and boundary conditions, as well as comparisons of computational cost and accuracy with established solvers, require further investigation.

## Data availability

[The GitHub repository for the implementation and benchmark scripts is:https://github.com/GRZpytorch/GNN_and_BEM]